# Recommendation system using a deep learning and graph analysis approach

Mahdi Kherad and Amir Jalaly Bidgoly

**Abstract**— When a user connects to the Internet to fulfill his needs, he often encounters a huge amount of related information. Recommender systems are the techniques for massively filtering information and offering the items that users find them satisfying and interesting. The advances in machine learning methods, especially deep learning, have led to great achievements in recommender systems, although these systems still suffer from challenges such as cold-start and sparsity problems. To solve these problems, context information such as user communication network is usually used. In this paper, we have proposed a novel recommendation method based on Matrix Factorization and graph analysis methods. In addition, we leverage deep Autoencoders to initialize users and items latent factors, and the deep embedding method gathers users' latent factors from the user trust graph. The proposed method is implemented on two standard datasets. The experimental results and comparisons demonstrate that the proposed approach is superior to the existing state-of-the-art recommendation methods. Our approach outperforms other comparative methods and achieves great improvements..

------------------------◆------------------------

## 1 INTRODUCTION

With the growing volume, complexity and dynamics of online information, the explosive growth of information available on the Internet often confuses users. The recommender systems are an effective key solution to overcome such information. These systems are useful information filtering tools to guide users in a personalized way for discovering the products or the services that may come from a wide range of possible options. Recommender systems play an important role in information systems to enhance business and facilitate decision making [1]. In general, the list of suggestions is based on user preferences, item features, past user interactions with items and some additional information such as temporal and spatial data. Recommender models are mainly classified into Collaborative Filtering (CF), Content-based, and hybrid recommender systems based on the types of input data [2]. However, these models have their limitations in dealing with cold start and data sparsity problems as well as the balance of suggestions quality in terms of different criteria [3-5].

The recommender system is an important part of the industry. It is a vital tool to promote sales and services for many online websites and mobile applications. For example, 80% of the videos watched on Netflix come from the recommender system [6] and 60% of video clicks in YouTube come from the home page suggestions [7]. By analyzing its user behavior, the recommender system proposes the most appropriate items (data, information, goods, etc.). This system is an approach designed to deal with the problems of large and growing volume of information and helps its user to reach their goal faster in the large volume of information [8]. In recommender systems, we try to identify and suggest the most appropriate item to suit the user's preference by guessing the user's thinking through information we have about his or her similar users and their opinions.

One of the recommended system design methods is the use of collaborative filtering based on Matrix Factorization (MF). This is a classic CF problem: Infer the missing entries in an $m \times$ n matrix, $R$, whose $(i, j)$ entry describes the ratings given by the $i$th user to the $j$th item. The performance is then measured using Root Mean Squared Error (RMSE)[9] . In this case, recommender systems serve as a two-way interaction between user interests and item features. MF as the most popular method of CF, it divides the rating matrix into two matrices of user and item latent factors with low-dimensions. MF can be considered as a predictive model learning process by estimating model factors from the training data (matrix R).

MF is an optimization problem in determining the model factors in order to best approximate the actual ratings with the prediction ratings. For an optimization problem, initialization is an important issue for the quality of the final solution. Common methods of MF initialize matrices of user and item factors based on quite simple mechanisms such as initializing by zero or random numbers. However, from an optimization perspective, MF-based methods are sensitive to the initialization of user and item factor matrices because the search space in the MF is non-convex. A suitable initialization can lead to a better local minimum and improve the efficiency and accuracy of the learning process [10].

To solve the cold start and data sparsity problems, context information is added to the recommender systems to enhance the predictor quality of the system. In particular, as the basis of interpersonal communication, the trust network between users plays an important role in solving information interaction, and experience communication. The basis of trust-based recommendations is based on the assumption that people usually prefer to make decisions based on the interests of their trusted friends rather than mass population. This assumption in addition to the analysis of social networks significantly creates algorithmic innovation, which can solve the above problems. Recently, trust-based recommender systems have attracted much attention [11] [12] [13]. The main issue in trust-based approaches is the use of MF technique to learn the latent features of users and items using the rating matrix and the user trust network.

Most existing trust-based models do not consider the diversity of user trust networks. This means users trust different friends on different topics. In fact, people tend to create different communities in a social network based on their social relationships,

• *M. kherad and A. Jalaly Bidgoly are with the Department of Computer Engineering University of Qom, Iran.*
*E-mail: M.Kherad@stu.qom.ac.ir, jalaly@qom.ac.ir.*

called community effects. Community structure is an important feature of social networks, and community detection has a significant impact on the discovery of social network structure. A community is defined as a set of nodes in a network that has more links than other nodes in the network. The purpose of community detection is to divide a graph into different subsets [14]. The contribution of recommendations from different friends' groups should be differentiated by their similarity with the target user. In other words, people in a community tend to trust each other and share common preferences with each other more than people in other communities. Therefore, community effect has a great impact on the performance of social recommendation methods.

On the other hand, the user's social status is of particular importance. The status is calculated differently in different environments and applications. A node with a high degree of centrality in the graph will be an effective node, regardless of the type of problem definition and environment. Nodes with a high social status perform a specific task and therefore need more attention. Identifying an important and influential node in the social network is an important challenge that requires defining precise criteria. Degree centrality, Betweenness centrality, Closeness centrality, PageRank and HITS are the most important methods for determining the status of a node in a network [15].

In addition, deep learning-based embedding techniques have demonstrated their power in many recommender tasks with the ability to extract representations from raw data. The application of embedding technique is not limited to images, texts and music, but also through the use of low-dimension embedding vectors, it provides an effective way to discover patterns of network structure [11].

In this paper, we propose a MF-based approach to a CF recommender system based on a combination of graph analysis and deep learning techniques. In the proposed method, we use network embedding technique [17] to learn the deep hidden information in users' trust network. We then use these embeddings to calculate trust and estimate user ratings so that the target user latent factors are more similar to the users they trust most. To reduce the time complexity, it is possible for each user, rather than calculating the similarity of all those in a community, to select the most important user of each community as the representative of that community, and to calculate their similarity to the target user. For this, in the proposed method, we apply the community detection technique [18] to the graph of users relationships. We then find the most effective and important node of each community [19] and incorporate it into the MF loss function to make the target user more similar to the most important node in their community. The proposed method also uses deep Autoencoder to initialize the MF latent factors of users and items.

The rest of this paper is organized as follows. In Section 2, a review of the literature on the recommender system is provided along with previous related works. Section 3 presents the details of the proposed method. The experimental results are reported in Section 4 and, the paper concludes in the last section.

## 2 LITERATURE REVIEW

The recommender system is used to estimate users' interests in items they have not seen [1]. There are mainly three types of recommender tasks based on output format namely rating predic-tion, ranking prediction (top-n recommendation) and classification. The goal of rating prediction is to fill in the missing elements of the user-item matrix. Ranking prediction creates a ranking list with $n$ items per user. The purpose of the classification task is to classify the candidate items into the correct categories for recommendations. In designing recommender systems, the aim is to improve the accuracy of predictions.

Recommender models are usually divided into three categories: collaborative filtering, content based and hybrid recommender system [1]. Collaborative filtering offers suggestions by learning from the user's historical interactions with the items, either explicitly (e.g., previous user rating) or implicit feedback (e.g., browser history). In this way, users' interests are predicted by analyzing the preferences of other users in the system and implicitly deduces similarities between them. Content-based recommender system is mainly based on a comparison between the auxiliary information given about the item and the users. A wide range of auxiliary information such as texts, images and videos can be taken into consideration. Content-based recommendations take into account contextual factors such as location, date, and time [20]. The hybrid model refers to a recommendation system that integrates two or more types of recommendation strategies [3].

As stated in [16], a recommender system contains a set of users $U = \{u_1, ..., u_m\}$ and a set of items $I = \{i_1, ..., i_n\}$. The ratings given by users to items are displayed in a rating matrix $R = [R_{u,i}]_{m \times n}$. In this matrix, $R_{u,i}$ represents the user rating of $u$ on item $i$. The social rating network can be represented as a graph that has two types of nodes i.e. user and item nodes. Edges between users indicate trust between two users, and edges between users and items indicate the rank that users give to items. $t_{u,v}$ indicates the value of social trust $u$ to $v$. The trust values are given in a matrix $[T_{u,v}]_{m \times m}$. Thus, in the trust-aware recommender system, user $u$ and item $i$ whose $R_{u,i}$ is unknown are given as inputs to predict user $u$ to item $i$ using $R$ and $T$.

Matrix Factorization (MF) [21] is a collaborative modeling technique and one of the most successful modeling methods. This method assumes that there are latent factors for users and items that are not in the data but that users are assigned items based on those factors. In MF, the ranking matrix $R$ is obtained by multiplying the two matrices $P \in R^{k \times m}$ as the user factor matrix and the $Q \in R^{k \times n}$ matrix of the item factors where $k$ is the number of latent factors. Determining the number of factors is an important issue in this type of learning. MF objective function is as follows:

$$L(R, P, Q) = \min_{P,Q} \frac{1}{2} \sum_{u=1}^{M} \sum_{i=1}^{N} I_{ui}(R_{u,i} - \hat{R}_{u,i}) + \frac{\lambda_P}{2} \|P\|_F^2 \\ + \frac{\lambda_Q}{2} \|Q\|_F^2 \quad (1)$$

Here, $\hat{R}_{u,i} = P_u^T Q_i$ is predicted rating of user $u$ on item $i$ and $I_{ui}$ is the indicator function that equals 1 if user $u$ rated item $i$ and equals 0 otherwise. $\lambda_P$ and $\lambda_Q$ are regularization terms to avoid model overfitting. $\|.\|_F^2$ denotes the Frobenius norm. The initial values of $P$ and $Q$ are always generated randomly or manually. Then, in each iteration, $P$ and $Q$ are updated by employing the stochastic gradient descent technique as follows:

$$P_u' = P_u - \alpha_1 \frac{\partial L}{\partial P_u} \quad (2)$$

$$Q'_i = Q_i - \alpha_1 \frac{\partial L}{\partial Q_i}$$

where $\alpha_1 > 0$ is the learning rate.

Mnih and Salakhutdinov [21] developed a Probabilistic Matrix Factorization (PMF) model to make predictions on large, sparsely and imbalanced Netflix dataset and perform better than other recommender methods. However, this traditional recommender method only uses rating history for the recommendation and ignores social relationship.

The relation of trust between users plays a key role in improving the quality of the recommender system. As online social networks services become more and more popular, the recommender system can obtain useful information from social networks [22]. Trust networks reflect trust relationships and the value of trust between users in social networks. Recently, various methods have been proposed for the recommendation system based on the use of the trust network among users. Moradi et al. [11] proposed a new reliability criterion based on the trust network to evaluate the effectiveness of the recommender system. They added this criterion to the trust-based collaborative filtering approach to improve the predictive quality of recommender systems based on social relationships. Ma et al. [12] present a trust relationship-based probabilistic graph algorithm, which incorporates the user's hobbies and the preferences of her/his trusted friends to optimize the MF objective function. Jamali and Ester [13], instead of just considering trust neighbors in calculating ratings, developed a random walk model that uses a combination of trust-based models and collaborative filtering, which allows us to define and measure the reliability of a recommendation. Jiang et al. [24] investigated various social recommender methods based on psychology and sociology research and proposed a method that adds social context information including user interests and influences between users to the MF model. Zheng et al.[25], to express the internal relations of social networks, propose a new hybrid model combining hypergraph theory with PMF. Pan et al. [26], to create a more accurate recommendation model based on the trust relationship between users, investigate the different roles that a user as a trustee and a trusted person in a social network, and present a new model of social MF based on adaptive trust network training to accurately reflect social relationships.

To overcome the sparsity of the user-item matrix, Massa and Avesani [27] used explicit user trust information to search for trusted users and recommend items that interest these users to the target user. Jamali and Ester [28] developed a recommender method for matrix factorization of social relations network called SocialMF and used the method of social relations propagation to improve the accuracy of the recommendation in the proposed method. They incorporated the trust propagation mechanism into the MF model where each user's factors depend on their direct neighbor factor vectors in the social network. Then, the latent factors of the user are generated by two components, the prior Gaussian distribution with a mean of zero to avoid over-fitting, and the conditional distribution of the latent attributes of the user according to the latent attributes of their neighbors.

Ma et al. [29] interpreted the differences between social - based and trust-aware recommender systems and proposed the Social Regularization based recommendation method (SoReg) to further improve traditional recommender systems. They used vector space similarity (VSS) and Pearson correlation coefficient (PCC) to measure the similarities of the two users and modeled the social network information as a regularization term to constrain the MF objective function. Tang et al. [30] utilized social relationships locally and globally for online recommender systems so that the global context can be gained by weighing the importance of user ratings based on user reputation scores.

The social status of a user indicates the importance of this user in the social relations network and indicates the level of communication of this user with others in the network. The social status of the user and homophily play an important role in improving the performance of the proposed systems [31]. The social status theory is used to explain how the different social levels of users influence the creation of trust relationships between users. Usually, a high-level user in the social network is regarded as an authority user, and a low-level user tends to create trust relation with a higher-level user. Yu et al. [22] propose an advanced matrix factorization model, considering the impact of users 'social status on users' trust relationships. Wang et al. [32] explore the prediction of trust relation in the sociological point of view and propose a new prediction algorithm of the user trust relation based on the study of the affect of social prestige and homophily theories on the trust relationship between users. Homophily indicates the tendency of persons to relationship to similar persons. Persons tend to interact with persons who are similar to themselves in a specified perspective [32].

Li and Ma [33] by considering the social status of the user and bias interest in creating a social relation, analyze the impact of social prestige on the user's social relationships and present a recommender approach based on the user's social relations. Tang et al. [24] investigated the effects of homophily on predicting the trust and integration of homophily in a MF model to optimize user latent factor space. Wang et al. [35] explore the impact of social status and homophily on trust and distrust, and propose a new method for predicting trust and distrust relations among users using multilayer neural networks and various factors namely homophily, emotion and status. Chen et al. [31] proposed a new social MF-based recommender system to improve recommendation quality by integrating user social status and homophily. They first build a network of user trust relationships based on user social relationships and rating information. The value of trust is then calculated using the trust propagation method and PageRank algorithm. Finally, trust relationships are integrated into the MF model to accurately predict the unknown ratings.

Deep learning is the use of artificial neural networks to perform learning tasks using multilayer networks. This technique has more learning power than neural networks [36]. Deep learning was first proposed by Hinton in 2006 [37]. Of course, the first steps of introducing and applying deep learning in the field of image processing, called convolutional neural networks, were performed by Lecun in 1998. In this method, the purpose was to perform multilevel learning and understanding of the image like the human brain [38]. Deep learning involves many techniques such as multilayer perceptron networks (MLP), Autoencoders (AE) [39], convolutional neural networks (CNN) [40], recurrent neural networks (RNNs) [41], deep belief networks (DBNs) [42]. Deep learning learns different levels of representation and abstraction of data that can solve supervised and unsupervised learning tasks [43].

The first attempts at using deep learning for recommender systems involved restricted Boltzmann machines (RBM) [44].

Several recent recommender approaches use AE [47] [50], MLP [47], and RNN [48]. Many popular MF techniques can be thought of as a form of dimension reduction. So it is natural that adapt deep Autoencoder s for this. Item-based Autoencoder (I-AutoRec) and user-based Autoencoder (U-AutoRec) are the first successful attempts to do so [50]. There are two ways to apply Autoencoder to the recommender system: 1) use Autoencoder to learn low-dimension features in the bottleneck layer and 2) fill in the blank entry of the ratings matrix in the reconstruction layer.

With the development of deep neural networks, distributed representation methods and embedding models have been extensively studied in recent years. Mikolov *et al.* [49] illustrated how to train the representations of words and phrases by the Skipgram model. Tang *et al.* [50] developed a new method for network embedding, which can easily work with networks of millions of nodes and edges. Since embedding methods can extract hierarchical representations of raw data, many researchers have also tried to incorporate these extracted factors into recommender systems. Zhao *et al.* [11], using network representation learning techniques, introduced a new approach to the recommendation task and it is cast into a similarity assessment process using embedding vectors. Liang *et al.* [51], inspired by the success of word embedding models, train item embeddings using a set of items that each user has rated. They propose an MF model to simultaneously decomposition the user-item interaction matrix and the item-to-item co-occurrence matrix with similar items factors. Zhao *et al.* [52] learned user and product features from an e-commerce website using recurring neural networks to apply knowledge extracted from social networking sites to products recommendation. They then developed a feature-based MF method using user embeddings for product recommendation. Guo *et al.* [53], using the advantages of network embedding techniques, proposed an embedding-based recommender approach consisting of embedding and collaborative filter models. They are first to use the hidden structure of social networks and rating patterns, a neural network-based embedding model pre-trained, which extracts user and item representations. Then, these extracted factors are combined in a collaborative filter model with linear hidden factors, which their method can not only use external information to improve the recommendation, but can also take advantage of the collaborative filtering techniques. Deng *et al.* [54] developed an MF-based approach for a trust-aware recommender system in social networks called Deep Learning based Matrix Factorization (DLMF). They examined the importance of initialization in MF and proposed a deep RBM-based initialization method. They then propose social trust ensemble learning model, which not only takes into account trusted friends' recommendations but also the effect of the community. In addition, they provide a community detection algorithm to find the community in a users' trust network.

# 3 THE PROPOSED FRAMEWORK

In this section, the proposed method for the recommender system using graph network analysis and deep learning is presented. The proposed framework combines community detection method, algorithm of find the most effective and important node, deep Autoencoder and deep embedding. The framework consists of seven steps. First, deep Autoencoder is used to initialize the latent factors of users and items in MF. The matrices of user and item factors are used to minimize the MF objective function using Gradient Descent algorithm. Then, using deep embedding method, user latent factors are extracted from the users trust network, which is used to calculate the value of trust and predict the user rating on the items. In the following, communities of users' trust network are identified, and the most important node in each community is found. The regularization term based on the most important node in each community is added to the MF function so that the features of each active user are more similar to the most important node in common community. Another regularization term added to MF function is based on trust values between users, which makes each user's preferences closer to the people they trust. The following subsections provide more details of the proposed method.

## 3.1 Initialization by deep Autoencoder

Given the non-convex objective function of MF, there is no guarantee that both factor matrices (P and Q) will be optimally determined [55]. In addition, matrix decomposition can converge to different local minimums with varying initial values of P and Q. Therefore, if the initial values are set correctly, the results will be closer to optimal than the situation where the initial values are set far from the global optimum. In this subsection, we explain how to use deep Autoencoder to pre-train the rating matrix and learn the initial values of latent attributes of users and items.

An Autoencoder is a neural network that implements two $encoder(x): R^n \rightarrow R^k$ and $decoder(x): R^k \rightarrow R^n$ mappings. The goal of Autoencoder is to obtain the k-dimensional representation of the data x so that the error measure between $x$ and $decode(encode(x))$ is minimized.

The proposed model uses a stack version of autocomplete and scaled exponential linear units (SELU) [56] and learn a deep architecture. The decoder architecture in the proposed model is symmetrical to the encoder and thus the number of parameters is halved. The purpose of deep Autoencoder (DAE) in the proposed method is to obtain user factors matrix ($P$) and item factors matrix ($Q$), thus using two deep Autoencoder : UDAE and IDAE. In fact, to obtain user factors ($P$), the inputs of UDAE are the rows of rating matrix ($R$) users and to obtain the item factors ($Q$), IDAE inputs are the rows of $Q$.

In DAE, both the encoder and decoder parts contain feedforward neural networks with $n$ fully connected layers to compute $f$ ($W * x + b$) where $f$ is a nonlinear activation function. The decoder weights $W_d^l$ correspond to the transverse weights of the encoder $W_e^l$ in layer $l$.

In the proposed method in UDAE for each input $x$, obtained latent factors ($z \in R^k$) considered as $P_u$ and in IDAE as $Q_i$. Since a zero vector as $x$ is not meaningful, we use the approach proposed by Sedhain et al. [50] and optimize Masked Mean Squared Error as loss function of DAE:

$$MMSE = \frac{m_i * (r_i - y_i)^2}{\sum_{i=0}^n m_i} \qquad (3)$$

where $r_i$ is actual rating, $y_i$ is reconstructed, or predicted rating, and $m_i$ is a mask function such that $m_i = 1\ if\ r_i \neq 0\ else\ m_i = 0$.

## 3.2 Users network embedding

In the proposed method, user factors are obtained by latent factors from rating matrix factorization and embedded factors obtained from pre-training the user's social network. Inspired by the success of the neural network-based embedding model in link prediction and node classification, the proposed method pretrains a network embedding model in [17] in a semi-supervised task and uses trained embeddings as user representations. node2vec can learn the representation of high-level stable features for nodes in any given network and obtain the diversity of connection patterns observed in networks with a random walk.

Consider a given network as $G = (A, E)$ where $A$ represents the set of nodes and $E$ represents the set of edges. For a node $m \in A$, let $N_S(m) \subset A$ be the network neighborhoods of node $m$ that is generated by $S$ strategy for neighborhoods sampling. Strategy $S$ is a random walk method that can detect neighbors by breadth-first or depth-first sampling. To learn the high-level representations of every node, node2vec tries to maximize the log-probability of observing neighbors $N_S(m)$ for node $m$ conditioned on its feature representations.

Since users in social networks often express their social interest through various friendships, a better understanding of these social networks is potentially useful for the recommender system. The type of social network can be a network of trust or friendship between users. Since node2vec output can be interpreted as high-level representations of network nodes, in the proposed method, we train node2vec to extract the deep social structure of the user trust network and consider $X_u \in \mathbb{R}^k$ as the extracted factors for the user $u$ from the user trust network. These extracted factors reflect the deep social interest of users. A linear combination of them shows the user how much will establish social links with others. This information can be useful for predicting ratings, especially when the users have rated very few items. Because social networks and ranting preferences potentially encode different types of information, combining them is expected to work best. A simple way to incorporate external factors into recommender systems is through a linear model, which means the sum of the latent factors of the collaborative filtering method with extracted social factors [53].

## 3.3 Calculate social trust

A trust network is a directed graph that nodes are users and edges are the trust relationship of a user to another. In this network, as the distances between users increase, the level of trust between users gradually decreases. As mentioned, people always prefer to trust their friends' recommendations because their friends' opinions are more reliable. But recommendations from trusted friends is not entirely appropriate for the target user, as they may have different habits, tastes, and preferences.

For users who are not directly connected, we use the trust multiplication calculation as the trust propagation operator. In addition, if there are several trust propagation paths, the shortest path is considered.

## 3.4 Community Detection

Users in social networks tend to form groups with high connections. These groups are called communities, clusters in different contexts. People in similar group tend to trust each other and share common preferences rather than with other groups. when comparing modularity optimization methods, speed and modularity value are two important criteria. Blondell et al. [18] showed that their method performs better than many similar modularity optimization methods in terms of modularity value and time complexity. Therefore, the proposed method uses the this method to identify communities of user trust network, which returns the community index for user $i$ as $c_i$.

Graph mining techniques are widely used for community detection in social networks because they are effective in identifying groups that are hidden in the data. This method for community discovery is a method for extracting communities from large networks. The algorithm is pretty fast and could be run with $O(n \log n)$ time complexity. This method is a greedy optimization method to maximize modularity value [28]. Modularity is a scale value between -1 to 1 that measures the density of edges within communities compared to edges between communities.

Theoretically, modularity optimization should result in the best grouping of nodes in a given network, but testing all possible states of nodes belonging to groups is impractical, so heuristic algorithms are used. The Louvain algorithm consists of repeated application of two steps. The first step is a greedy assignment of nodes to communities, favoring local optimizations of modularity. The second step is the definition of a new coarse-grained network based on the communities found in the first step. These two steps are repeated until no further modularity-increasing reassignments of communities are possible.

## 3.5 Most important nodes

In social networks, users with high social status usually provide more valuable information than users with low social status. These users are sometimes referred to as opinion leaders since they have a great impact on other users' opinions. Identify the most influential people in the network from the perspective of various parameters can find the nodes that need more attention and investment to perform a specific task [60]. Obviously, the importance of different people in a community is not the same. Some are more important because of their social status, relationships or friends with their influence. Therefore, some of these criteria not only matter to the number of friends of each person, but also to the network of friends of each person's friends and the network of more mediated friends. In trust social network, there may be nodes that are more trusted by people and more likely to accept their experiences and opinions. In social network analysis, graph-based metrics and common heuristic methods are used to identify influential nodes in social networks such as Degree Centrality, Distance Centrality, Closeness Centrality, and Betweenness Centrality.

In our proposed method, the graph-based metric is used as a measure of finding the most effective user in ever community of trust network and the node that has the highest value of score in each community is used to influence the rating of other users in that community.

## 3.6 Regularization terms

Intuitively, users tend to share similar preferences about items with their trusted friends in the shared community. On the other hand, in any community, a node with a high social status has the most impact on other people, so other people in the community tend to be like the most effective person in the community. In addition, independent of the community in which users are present, the similarity of users depends on the value of trust between users, that is, the greater the trust value between users, the more similar they are to each other.

Based on the above intuitions, in the proposed method we add two regularization terms to the recommender model and modify the MF problem (Equation 1).

To find the local optimal of Equation 8, we used the stochastic gradient descent algorithm and update the latent factors $P$, $Q$ and weights $W$ with the gradients.

# 4 EXPERIMENTAL RESULTS

Ciao and Epinions standard datasets used in this paper have been published publicly by Tang et al. [۳۰], which includes user ratings on items, rating time, and social network between users. Because the rating matrix is very sparse, the recommender problem in these two datasets is challenging.

Ciao dataset is collected from the online comments site *www.ciao.com*, a multi-million-user online community that provides a forum for registered users to write their own opinions on a wide range of products to help others make better decisions. The Epinions dataset is from a former popular website (Epinions.com) for product reviews, launched in 1999. At Epinions, visitors are allowed to read other users' comments about the products and services to help make purchasing decisions. While both websites are now formally closed, but their dataset is available for academic research. In both datasets, registered users express their opinions by rating the product or service using an integer from 1 to 5 and provide a trust list to determine in which order the product views are shown to visitors. Table 1 shows the statistics for Ciao and Epinions datasets.

TABLE 1
STATISTICS OF CIAO AND EPINIONS DATASETS

| dataset | #users | #items | #ratings | Rating density | #trust relations |
|---|---|---|---|---|---|
| Epinions | 49290 | 139738 | 664824 | 0.00009 | 478181 |
| Ciao | 7375 | 106797 | 284086 | 0.0004 | 111781 |

In this paper, the root mean square error (RMSE) [۱۲] is used to evaluate the performance of the proposed method, which is defined as follows:

$$RMSE = \sqrt{\frac{\sum_{u,i}(R_{u,i} - \hat{R}_{u,i})^2}{N}} \quad (10)$$

Where $R_{u,i}$ is real and $\hat{R}_{u,i}$ is predicted rating of user $u$ on item $i$ and $N$ is the number of ratings used for the test. Since RMSE measures the prediction error of the recommender method, the lower value of the RMSE indicates that a method can predict more accurately.

To implement and run of proposed model, Python 3.7 was used in Spyder environment on a computer with 8 GB of RAM and a 2.2 GHz quad-core processor. The Keras [۶۱] library in Python is used to implement deep Autoencoder neural networks. The hyper-parameters of the proposed method for both datasets are set in Table 2.

TABLE 2
THE HYPER-PARAMETER VALUES OF PROPOSED METHOD

| Symbol | Description | Value |
|---|---|---|
| $k$ | The dimension of latent features | 10 |
| $\alpha_1$ | The learning rate of MF | 0.005 |
| $\alpha_2$ | The learning rate of DAE | 0.001 |
| $\lambda_w$ | The regularization parameter of weights of extracted user factors from node2vec | 0.1 |
| $\lambda_P$ | The regularization constant of user latent factors | 0.1 |
| $\lambda_Q$ | The regularization constant of item latent factors | 0.1 |
| $\lambda_T$ | The tradeoff parameter plays the role of adjusting the effects of interpersonal trust between users | 0.1 |
| $\lambda_C$ | The control parameter for the effect of the most important user in community | 0.1 |
| $bs$ | The batch size | 128 |
| $n_L$ | The number of the hidden layers of DAE | 7 |
| $L_1$ | The number of neurons in the hidden layer 1 | 128 |
| $L_2$ | The number of neurons in the hidden layer 2 | 64 |
| $L_3$ | The number of neurons in the hidden layer 3 | 32 |
| $L_4$ | The number of neurons in the hidden layer 4 | 10 |
| $L_5$ | The number of neurons in the hidden layer 5 | 32 |
| $L_6$ | The number of neurons in the hidden layer 6 | 64 |
| $L_7$ | The number of neurons in the hidden layer 7 | 128 |

In order to evaluate the performance of the proposed method, it is compared with the following state-of-the-art RS methods:

- PMF [۲۲]: **P**robabilistic **M**atrix **F**actorization is a basic recommendation method which seeks to approximate the rating matrix by multiplication of lower rank factors. In this method, only rating data is used, and models the latent factors of users and items with the Gaussian distribution.
- SoRec [۱۶]: **So**cial **Rec**ommendation method performs co-factorization the user-item ranking matrix and the user-user social relations matrix.
- SoReg [۱۹]: **So**cial **Reg**ularization is another popular recommendation method that model social network information as social regularization terms to constrain the MF objective function.
- SocialMF [۲۸]: This method adds trust information and trust propagation to MF model for recommender systems.
- TrustMF [۶۳]: This method adopts MF technique to map users into two low-dimensional spaces, truster space and trustee space, by factorization trust network according to the trust directional property.
- NeuMF [۴۷]: This method is a state-of-the-art MF model with neural network architecture. The original version is for recommendations ranking task but has been modified its loss function to rating prediction.
- DeepSoR [۱۴]: **Deep** Modeling of **So**cial **R**elations for recommendation method uses a deep neural network to learn the representations of each user from social relationships that integrate with PMF to predict ratings.
- GCMC [۱۵]: **G**raph **C**onvolutional **M**atrix **C**ompletion method is a graph Autoencoder framework, which creates hidden features of users and items through a differentiable message passing on the user-item graph.
- MFn2v+ [۵۲]: It uses a network embedding model to learn representations of users from a social network and items from a sequence of items, and integrates the trained embeddings into the factors of MF model linearly.

- GraphRec [٦٦]: A novel graph neural network framework for social recommendations that can model graph data in social recommendations while simultaneously incorporating interactions and opinions into the user-item graph.

In this paper, RMSE of the above methods obtained in the experiments in [65] [51] on Ciao and Epinions datasets are used to compare with the proposed method results. In these experiments, the datasets are split into two parts of 80% and 20%, for training and testing, respectively. In order to be able to compare, the same ratio of division is considered in the proposed method. Also the parameters of the state-of-the-art algorithms are set as specified in the corresponding papers with optimal performance. Table 3 shows the RMSE of rating prediction in the proposed method and state-of-the-art RS methods in Ciao and Epinions data sets.

TABLE 3
RMSE COMPARISON OF DIFFERENT RECOMMENDER SYSTEMS

| Method/dataset | Epinions | Ciao |
|---|---|---|
| PMF | ۱,۲۱۲۸ | ۱,۱۲۳۸ |
| SoRec | ۱,۱۴۳۷ | ۱,۰۵۲ |
| SoReg | ۱,۱۷۰۳ | ۱,۰۸٤۸ |
| SocialMF | ۱,۱۳۲۸ | ۱,۰۵۰۱ |
| TrustMF | ۱,۱۳۹۵ | ۱,۰٤۷۹ |
| NeuMF | ۱,۱٤۷۶ | ۱,۰٦۱۷ |
| DeepSoR | ۱,۰۹۷۲ | ۱,۰۳۱٦ |
| GCMC | ۱,۰۷۱۱ | ۰,۹۹۳۱ |
| GraphRec | ۱,۰۷۳۱ | ۰,۹۷۹٤ |
| MFn2v+ | ۱,۰٤۱ | ۰,۹۵۷ |
| **Proposed method** | ۰,۹۰۰۸ | ۰,۸۰۸۲ |

As can be seen in Table 3, the proposed method has achieved lower RMSE than the other methods in both datasets. From these results, it can be seen that since the PMF only uses ranking matrix information for recommendations, it performs worse than the other methods in both datasets. Whereas, TrustMF, SocialMF, SoRec, SoReg methods, which also use social network information of users, achieve better results than PMF. It can be concluded that it is necessary to consider the social network of users in order to achieve more accurate results in recommendation systems. On the other hand, the results of deep learning based methods (NeuMF) are also better than the PMF and comparable to the social network based methods (TrustMF, SocialMF, SoRec, SoReg) which can be concluded that the deep neural network model also improves the recommendations. DeepSoR, GCMC, GraphRec, and MFn2v+ approaches that take advantages of users' social network information alongside deep learning power work better than the ones which only use either social network information or deep learning methods. Among these methods, GraphRec and MFn2v+ show strong performance. This means that deep embedding is useful in learning representation for graph data, because it naturally integrates node information as well as topological structure. The proposed method, not only take the advantages of all of the above approaches but also by employing social graph analysis techniques, has made its results superior to all other state-of-the-art methods.

## 5 CONCLUSIONS

In this paper, a collaborative filtering recommender method based on matrix factorization is proposed that first initializes latent factors of users and items by deep Autoencoder. It then captures user representations using user trust network embedding. These extracted representations of users are used to calculate users' trust and predict ratings of users on items.

Users in social networks pay more attention to the opinions of people they trust than others. Besides, users are more likely to be connected with the ones who have similar interests, hence users of a community are more likely to have similar views. On the other hand, in a socoiety, users are interested in following and imitating the opinion of important people in the community. In the proposed method, according to these intuitions, regularization terms are added to the objective function of MF so that user's interests become similar to those of the trusted user and the most effective person in the community. We use community detection and algorithm to find the most important user in ever community.

The RMSE results of the proposed method on standard two datasets compared to the state-of-the-art recommender methods show the superiority of the proposed method. By comparing different methods, it can be concluded that using information of social network and deep neural networks along with rating matrix information empowers the recommender methods.

Although the proposed method has advantages, there are still limitations to this model. The proposed method utilizes user network information, in future work the network information of the items and other features of the users and items such as user conditions, geographical location and time of ratings of users and items in the recommender system can be used. In addition, the social network of users and items can be used dynamically.

## REFERENCES

[1] S. Zhang, L. Yao, A. Sun, and Y. Tay, "Deep learning based recommender system: A survey and new perspectives," *ACM Computing Surveys (CSUR),* vol. 52, no. 1, p. 5, 2019.

[2] G. Adomavicius and A.Tuzhilin, "Toward the next generation of recommender systems: A survey of the state-of-the-art and possible extensions.," *IEEE Transactions on Knowledge and Data Engineering,* vol. 17, no. 6, pp. 734-749, 2005.

[3] R.Burke, "Hybrid recommender systems: Survey and experiments.," *User Modeling and User Adapted Interaction,* vol. 12, no. 4, pp. 331-370, 2002.

[4] S. M. McNee, J. Riedl, and J. A. Konstan, "Being accurate is not enough: how accuracy metrics have hurt recommender systems," in *CHI'06 extended abstracts on Human factors in computing systems,* 2006, pp. 1097-1101: ACM.

[5] S. Vargas and P. Castells, "Rank and relevance in novelty and diversity metrics for recommender systems," in *Proceedings of the fifth ACM conference on Recommender systems,* 2011, pp. 109-116: ACM.

[6] C. A. Gomez-Uribe and N. Hunt, "The netflix recommender system: Algorithms, business value, and innovation," *ACM Transactions on Management Information Systems (TMIS),* vol. 6, no. 4, p. 13, 2016.

[7] J. Davidson *et al.*, "The YouTube video recommendation system," in *Proceedings of the fourth ACM conference on Recommender systems,* 2010, pp. 293-296: ACM.

[8] F. Ricci, L. Rokach, and B. Shapira, "Introduction to Recommender Systems Handbook," in *Recommender Systems*

*Handbook*: Springer, 2011.

[9] O. Kuchaiev and B. Ginsburg, "Training deep autoencoders for collaborative filtering," *arXiv preprint arXiv:1708.01715,* 2017.

[10] R. Zdunek, "Initialization of nonnegative matrix factorization with vertices of convex polytope," in *International Conference on Artificial Intelligence and Soft Computing*, 2012, pp. 448-455: Springer.

[11] P. Moradi and S. Ahmadian, "A reliability-based recommendation method to improve trust-aware recommender systems," *Expert Systems with Applications,* vol. 42, no. 21, pp. 7386-7398, 2015.

[12] H. Ma, I. King, and M. R. Lyu, "Learning to recommend with social trust ensemble," in *Proceedings of the 32nd international ACM SIGIR conference on Research and development in information retrieval*, 2009, pp. 203-210.

[13] M. Jamali and M. Ester, "Trustwalker: a random walk model for combining trust-based and item-based recommendation," in *Proceedings of the 15th ACM SIGKDD international conference on Knowledge discovery and data mining*, 2009, pp. 397-406.

[14] C. Wang, W. Tang, B. Sun, J. Fang, and Y. Wang, "Review on community detection algorithms in social networks," in *Progress in Informatics and Computing (PIC), 2015 IEEE International Conference on*, 2015, pp. 551-555: IEEE.

[15] M. Newman, *Networks*. Oxford university press, 2018.

[16] W. X. Zhao, J. Huang, and J.-R. Wen, "Learning distributed representations for recommender systems with a network embedding approach," in *Asia Information Retrieval Symposium*, 2016, pp. 224-236: Springer.

[17] A. Grover and J. Leskovec, "node2vec: Scalable feature learning for networks," in *Proceedings of the 22nd ACM SIGKDD international conference on Knowledge discovery and data mining*, 2016, pp. 855-864.

[18] V. D. Blondel, J.-L. Guillaume, R. Lambiotte, and E. Lefebvre, "Fast unfolding of communities in large networks," *Journal of statistical mechanics: theory and experiment,* vol. 2008, no. 10, p. P10008, 2008.

[19] J. M. Kleinberg, "Hubs, authorities, and communities," *ACM computing surveys (CSUR),* vol. 31, no. 4es, pp. 5-es, 1999.

[20] G. Adomavicius and A. Tuzhilin, "Context-aware recommender systems," in *Recommender systems handbook*: Springer, 2011, pp. 217-253.

[21] N. Guan, D. Tao, Z. Luo, and B. Yuan, "Online nonnegative matrix factorization with robust stochastic approximation," *IEEE Transactions on Neural Networks and Learning Systems,* vol. 23, no. 7, pp. 1087-1099, 2012.

[22] A. Mnih and R. R. Salakhutdinov, "Probabilistic matrix factorization," in *Advances in neural information processing systems*, 2008, pp. 1257-1264.

[23] Y. Yu, Y. Gao, H. Wang, and S. Sun, "Integrating user social status and matrix factorization for item recommendation," *J. Comput. Res. Develop.,* vol. 55, no. 1, pp. 113-124, 2018.

[24] M. Jiang *et al.*, "Social contextual recommendation," in *Proceedings of the 21st ACM international conference on Information and knowledge management*, 2012, pp. 45-54.

[25] X. Zheng, Y. Luo, L. Sun, X. Ding, and J. Zhang, "A novel social network hybrid recommender system based on hypergraph topologic structure," *World Wide Web,* vol. 21, no. 4, pp. 985-1013, 2018.

[26] Y. Pan, F. He, and H. Yu, "Social recommendation algorithm using

[27] P. Massa and P. Avesani, "Trust-aware recommender systems," in *Proceedings of the 2007 ACM conference on Recommender systems*, 2007, pp. 17-24.

[28] M. Jamali and M. Ester, "A matrix factorization technique with trust propagation for recommendation in social networks," in *Proceedings of the fourth ACM conference on Recommender systems*, 2010, pp. 135-142.

[29] H. Ma, D. Zhou, C. Liu, M. R. Lyu, and I. King, "Recommender systems with social regularization," in *Proceedings of the fourth ACM international conference on Web search and data mining*, 2011, pp. 287-296.

[30] J. Tang, X. Hu, H. Gao, and H. Liu, "Exploiting local and global social context for recommendation," in *Twenty-Third International Joint Conference on Artificial Intelligence*, 2013.

[31] R. Chen *et al.*, "A novel social recommendation method fusing user's social status and homophily based on matrix factorization techniques," *IEEE Access,* vol. 7, pp. 18783-18798, 2019.

[32] Y. Wang, X. Wang, and W. Zuo, "Trust prediction modeling based on social theories," *J. Softw.,* vol. 25, no. 12, pp. 2893-2904, 2014.

[33] H. Li, X.-P. Ma, and J. Shi, "Incorporating trust relation with PMF to enhance social network recommendation performance," *International Journal of Pattern Recognition and Artificial Intelligence,* vol. 30, no. 06, p. 1659016, 2016.

[34] J. Tang, H. Gao, X. Hu, and H. Liu, "Exploiting homophily effect for trust prediction," in *Proceedings of the sixth ACM international conference on Web search and data mining*, 2013, pp. 53-62.

[35] X. Wang, Y. Wang, and H. Sun, "Exploring the combination of Dempster-Shafer theory and neural network for predicting trust and distrust," *Computational intelligence and neuroscience,* vol. 2016, 2016.

[36] L. Zhang, S. Wang, and B. Liu, "Deep learning for sentiment analysis: A survey," *Wiley Interdisciplinary Reviews: Data Mining and Knowledge Discovery,* p. e1253, 2018.

[37] G. E. Hinton, N. Srivastava, A. Krizhevsky, I. Sutskever, and R. R. Salakhutdinov, "Improving neural networks by preventing co-adaptation of feature detectors," *arXiv preprint arXiv:1207.0580,* 2012.

[38] Y. LeCun, L. Bottou, Y. Bengio, and P. Haffner, "Gradient-based learning applied to document recognition," *Proceedings of the IEEE,* vol. 86, no. 11, pp. 2278-2324, 1998.

[39] M. Chen, Z. Xu, K. Weinberger, and F. Sha, "Marginalized denoising autoencoders for domain adaptation," *arXiv preprint arXiv:1206.4683,* 2012.

[40] I. Goodfellow, Y. Bengio, and A. Courville, *Deep learning*. MIT press, 2016.

[41] Y. Bengio, I. J. Goodfellow, and A. Courville, "Deep learning," *Nature,* vol. 521, no. 7553, pp. 436-444, 2015.

[42] Q. T. Ain *et al.*, "Sentiment analysis using deep learning techniques: a review," *Int J Adv Comput Sci Appl,* vol. 8, no. 6, p. 424, 2017.

[43] L. Deng and D. Yu, "Deep learning: methods and applications," *Foundations and Trends® in Signal Processing,* vol. 7, no. 3–4, pp. 197-387, 2014.

[44] R. Salakhutdinov, A. Mnih, and G. Hinton, "Restricted Boltzmann machines for collaborative filtering," in *Proceedings of the 24th

*international conference on Machine learning*, 2007, pp. 791-798: ACM.

[45] S. Sedhain, A. K. Menon, S. Sanner, and L. Xie, "Autorec: Autoencoders meet collaborative filtering," in *Proceedings of the 24th International Conference on World Wide Web*, 2015, pp. 111-112: ACM.

[46] F. Strub and J. Mary, "Collaborative filtering with stacked denoising autoencoders and sparse inputs," in *NIPS workshop on machine learning for eCommerce*, 2015.

[47] X. He, L. Liao, H. Zhang, L. Nie, X. Hu, and T.-S. Chua, "Neural collaborative filtering," in *Proceedings of the 26th International Conference on World Wide Web*, 2017, pp. 173-182: International World Wide Web Conferences Steering Committee.

[48] C.-Y. Wu, A. Ahmed, A. Beutel, A. J. Smola, and H. Jing, "Recurrent recommender networks," in *Proceedings of the tenth ACM international conference on web search and data mining*, 2017, pp. 495-503: ACM.

[49] T. Mikolov, I. Sutskever, K. Chen, G. S. Corrado, and J. Dean, "Distributed representations of words and phrases and their compositionality," in *Advances in neural information processing systems*, 2013, pp. 3111-3119.

[50] J. Tang, M. Qu, M. Wang, M. Zhang, J. Yan, and Q. Mei, "Line: Large-scale information network embedding," in *Proceedings of the 24th international conference on world wide web*, 2015, pp. 1067-1077.

[51] D. Liang, J. Altosaar, L. Charlin, and D. M. Blei, "Factorization meets the item embedding: Regularizing matrix factorization with item co-occurrence," in *Proceedings of the 10th ACM conference on recommender systems*, 2016, pp. 59-66.

[52] W. X. Zhao, S. Li, Y. He, E. Y. Chang, J.-R. Wen, and X. Li, "Connecting social media to e-commerce: Cold-start product recommendation using microblogging information," *IEEE Transactions on Knowledge and Data Engineering,* vol. 28, no. 5, pp. 1147-1159, 2015.

[53] L. Guo, Y.-F. Wen, and X.-H. Wang, "Exploiting pre-trained network embeddings for recommendations in social networks," *Journal of Computer Science and Technology,* vol. 33, no. 4, pp. 682-696, 2018.

[54] S. Deng, L. Huang, G. Xu, X. Wu, and Z. Wu, "On deep learning for trust-aware recommendations in social networks," *IEEE transactions on neural networks and learning systems,* vol. 28, no. 5, pp. 1164-1177, 2016.

[55] M. Rezaei, R. Boostani, and M. Rezaei, "An efficient initialization method for nonnegative matrix factorization," *Journal of Applied Sciences,* vol. 11, no. 2, pp. 354-359, 2011.

[56] G. Klambauer, T. Unterthiner, A. Mayr, and S. Hochreiter, "Self-normalizing neural networks," in *Advances in neural information processing systems*, 2017, pp. 971-980.

[57] R. Kannan, M. Ishteva, and H. Park, "Bounded matrix factorization for recommender system," *Knowledge and information systems,* vol. 39, no. 3, pp. 491-511, 2014.

[58] A. Lancichinetti and S. Fortunato, "Community detection algorithms: a comparative analysis," *Physical review E,* vol. 80, no. 5, p. 056117, 2009.

[59] T. Aynaud and J. Guillaume, "Static community detection algorithms for evolving networks," in *8th International Symposium on Modeling and Optimization in Mobile, Ad Hoc, and Wireless Networks*, 2010, pp. 513-519.

[60] J. Zhou, Y. Zhang, and J. Cheng, "Preference-based mining of top-K influential nodes in social networks," *Future Generation Computer Systems,* vol. 31, pp. 40-47, 2014.

[61] F. Chollet, "keras, GitHub," *GitHub repository, https://github. com/fchollet/keras,* 2015.

[62] H. Ma, H. Yang, M. R. Lyu, and I. King, "Sorec: social recommendation using probabilistic matrix factorization," in *Proceedings of the 17th ACM conference on Information and knowledge management*, 2008, pp. 931-940.

[63] B. Yang, Y. Lei, J. Liu, and W. Li, "Social collaborative filtering by trust," *IEEE transactions on pattern analysis and machine intelligence,* vol. 39, no. 8, pp. 1633-1647, 2016.

[64] W. Fan, Q. Li, and M. Cheng, "Deep modeling of social relations for recommendation," in *Thirty-Second AAAI Conference on Artificial Intelligence*, 2018.

[65] R. v. d. Berg, T. N. Kipf, and M. Welling, "Graph convolutional matrix completion," *arXiv preprint arXiv:1706.02263,* 2017.

[66] W. Fan *et al.*, "Graph neural networks for social recommendation," in *The World Wide Web Conference*, 2019, pp. 417-426.

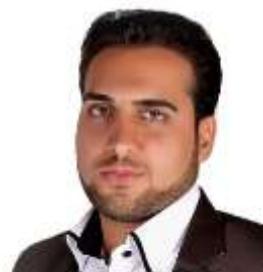

**Mahdi Kherad** received his Eng. degree in computer engineering from the University of Birjand, Iran in 2012 and his M.Sc. degree in Information Technology from University of Birjand, Iran in 2016. He is currently a Ph.D. student at the Technology and Engineering Department, University of Qom, Iran. His research interests include machine learning, deep learning, and data mining.

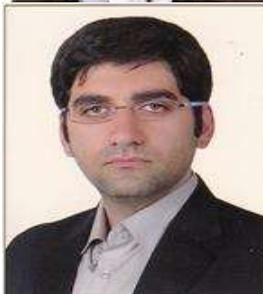

**Amir Jalaly Bidgoly** received the Ph.D. degree in computer engineering from Isfahan University, Isfahan, Iran, in 2015.,He is currently an Associate Professor with the Department of Computer Engineering, University of Qom. His research interests include system security, social security, trust and reputation systems, and formal verification..